\def\year{2018}\relax
\documentclass[letterpaper]{article} 
\usepackage{aaai18}  
\usepackage{times}  
\usepackage{helvet}  
\usepackage{courier}  
\usepackage{url}  
\usepackage{graphicx}  
\usepackage[utf8]{inputenc} 
\usepackage[T1]{fontenc}    
\usepackage{url}            
\usepackage{booktabs}       
\usepackage{amsfonts}       
\usepackage{nicefrac}       
\usepackage{microtype}      
\usepackage{amsmath}
\usepackage{algorithm}
\usepackage{multirow}
\usepackage{rotating}
\usepackage{graphicx}
\usepackage{bbm}
\usepackage{color}
\usepackage{placeins}
\usepackage{wrapfig}
\usepackage{sidecap}
\usepackage{booktabs}
\usepackage[noend]{algpseudocode}
\algnewcommand{\Initialize}[1]{%
  \State \textbf{Initialize:}
  \Statex \hspace*{\algorithmicindent}\parbox[t]{.8\linewidth}{\raggedright #1}
}
\algnewcommand{\Inputs}[1]{%
  \State \textbf{Inputs:}
  \Statex \hspace*{\algorithmicindent}\parbox[t]{.8\linewidth}{\raggedright #1}
}
\algnewcommand{\Outputs}[1]{%
  \State \textbf{Outputs:}
  \Statex \hspace*{\algorithmicindent}\parbox[t]{.8\linewidth}{\raggedright #1}
}

\begin{document}
%
\title{A Continuous Relaxation of Beam Search for End-to-end Training of Neural Sequence Models}
\author{Kartik Goyal\\
Carnegie Mellon University\\
\texttt{kartikgo@cs.cmu.edu}\\
\And
  Graham Neubig\\
  Carnegie Mellon University \\
  \texttt{gneubig@cs.cmu.edu} \\
  \And
  Chris Dyer\\
  Deepmind\\
  \texttt{cdyer@google.com} \\
  \And
  Taylor Berg-Kirkpatrick\\
  Carnegie Mellon University\\
  \texttt{tberg@cs.cmu.edu} \\
}
\maketitle
\begin{abstract}
\textit{Beam search} is a desirable choice of test-time decoding algorithm for neural sequence models because it potentially avoids search errors made by simpler greedy methods. However, typical cross entropy training procedures for these models do not directly consider the behaviour of the final decoding method. As a result, for cross-entropy trained models, beam decoding can sometimes yield reduced test performance when compared with greedy decoding. In order to train models that can more effectively make use of beam search, we propose a new training procedure that focuses on the final loss metric (e.g. Hamming loss) evaluated on the output of beam search. While well-defined, this ``direct loss'' objective is itself discontinuous and thus difficult to optimize. Hence, in our approach, we form a sub-differentiable surrogate objective by introducing a novel continuous approximation of the beam search decoding procedure. In experiments, we show that optimizing this new training objective yields substantially better results on two sequence tasks (Named Entity Recognition and CCG Supertagging) when compared with both cross entropy trained greedy decoding and cross entropy trained beam decoding baselines.
\end{abstract}
\section{Introduction}
\begin{algorithm*}[t]
\caption{Standard Beam Search}\label{hard}
\begin{algorithmic}[1]
\Initialize{\strut$h_{0,i} \gets \vec{0}$, $e_{0,i} \gets \textit{embedding(<s>)}$, $s_{0,i} \gets 0$, $i=1,\ldots,k$ 
}
\For{t = 0 to T}
\For{i = 1 to k}
\ForAll{$v \in V$}
\State $\tilde{s}_t[i,v] \gets s_{t,i} + f(h_{t,i},v)$ \Comment{$f$ is the local output scoring function}
\EndFor
\EndFor
 \State $s_{t+1} \gets$ \textit{top-k-max}$(\tilde{s}_t)$ \Comment{Top \textit{k} values of the input matrix}
 \State $\textit{b}_{t+1, *}, y_{t, *} \gets$ \textit{top-k-argmax}$(\tilde{s}_t)$ \Comment{Top $k$ argmax index pairs of the input matrix}
 \For{i = 1 to k}
  \State $e_{t+1,i} \gets$ \textit{embedding}($y_{t,i}$)
  \State $h_{t+1,i} \gets r(h_{t,i},e_{t+1,i})$ \Comment{$r$ is a nonlinear recurrent function that returns state at next step}
 \EndFor
\EndFor
 \State $\hat{y}$ $\gets$ \textit{follow-backpointer}($(b_{1,*},y_{1,*}),\ldots, (b_{T,*},y_{T,*})$)
 \State $s(\hat{y}) \gets \textit{max}(s_T)$
\end{algorithmic}
\end{algorithm*}
Sequence-to-sequence (seq2seq) models have been successfully used for many sequential decision tasks such as machine translation \cite{sutskever2014sequence,bahdanau2014neural}, parsing \cite{dyer2016recurrent,dyer2015transition}, summarization \cite{rush2015neural}, dialog generation \cite{serban2015building}, and image captioning \cite{xu2015show}.
\textit{Beam search} is a desirable choice of test-time decoding algorithm for such models because it potentially avoids search errors made by simpler greedy methods. However, the typical approach to training neural sequence models is to use a locally normalized maximum likelihood objective (cross-entropy training) \cite{sutskever2014sequence}. This objective does not directly reason about the behaviour of the final decoding method. As a result, for cross-entropy trained models, beam decoding can sometimes yield reduced test performance when compared with greedy decoding \cite{koehn2017six,neubig2017neural,cho2014properties}.
These negative results are not unexpected. The training procedure was not \emph{search-aware}: it was not able to consider the effect that changing the model's scores might have on the ease of search while using a beam decoding, greedy decoding, or otherwise. 

We hypothesize that the under-performance of beam search in certain scenarios can be resolved by using a better designed training objective.
Because beam search potentially offers more accurate search when compared to greedy decoding, we hope that appropriately trained models should be able to leverage beam search to improve performance. In order to train models that can more effectively make use of beam search, we propose a new training procedure that focuses on the final loss metric (e.g. Hamming loss) evaluated on the output of beam search. While well-defined and a valid training criterion, this ``direct loss'' objective is discontinuous and thus difficult to optimize. Hence, in our approach, we form a sub-differentiable surrogate objective by introducing a novel continuous approximation of the beam search decoding procedure. In experiments, we show that optimizing this new training objective yields substantially better results on two sequence tasks (Named Entity Recognition and CCG Supertagging) when compared with both cross-entropy trained greedy decoding and cross-entropy trained beam decoding baselines. 

Several related methods, including  reinforcement learning \cite{ranzato2015sequence,bahdanau2016actor}, imitation learning \cite{daume2009search,ross2011reduction,bengioss}, and discrete search based methods \cite{wiseman2016sequence,andor2016globally,daume2005learning,gormley2015approximation}, have also been proposed to make training search-aware.
These methods include approaches that forgo direct optimization of a global training objective, instead incorporating credit assignment for search errors by using methods like \textit{early updates} \cite{collins2004incremental} that explicitly track the reachability of the gold target sequence during the search procedure.  While addressing a related problem -- credit assignment for search errors during training -- in this paper, we propose an approach with a novel property: we directly optimize a continuous and global training objective using backpropagation. As a result, in our approach, credit assignment is handled directly via gradient optimization in an end-to-end computation graph. The most closely related work to our own approach was proposed by Goyal et al. \cite{softgreedy}. They do not consider beam search, but develop a continuous approximation of greedy decoding for scheduled sampling objectives. Other related work involves training a generator with a Gumbel reparamterized sampling module to more reliably find the MAP sequences at decode-time \cite{gu2017neural}, and constructing surrogate loss functions \cite{bahdanau2016task} that are close to task losses.

\section{Model}

We denote the seq2seq model parameterized by $\theta$ as $\mathcal{M}(\theta)$. We denote the input sequence as $x$, the gold output sequence as $y^*$ and the result of beam search over $\mathcal{M}(\theta)$ as $\hat{y} = \textit{Beam}(x,\mathcal{M}(\theta))$. Ideally, we would like to directly minimize a final evaluation loss, $L(\hat{y},y^*)$, evaluated on the result of running beam search with input $x$ and model $\mathcal{M}(\theta)$. Throughout this paper we assume that the evaluation loss decomposes over time steps $t$ as: $L(\hat{y},y^*)=\sum_{t=1}^T d(\hat{y}_t,y^*)$\footnote{This assumption does not hold  for some popular evaluation metrics (e.g. BLEU). In these cases, surrogate evaluation losses such as Hamming distance can be used
.}. 
We refer to this idealized training objective that directly evaluates prediction loss as the ``direct loss'' objective and define it as:
\begin{align}
\min_{\theta} G_{\textrm{DL}}(x,\theta,y^*) = \min_{\theta} L(\textit{Beam}(x,\mathcal{M}(\theta)),y^*)
\end{align}
Unfortunately, optimizing this objective using gradient methods is difficult because the objective is discontinuous. The two sources of discontinuity are:
\begin{enumerate}
\item As we describe later in more detail, beam search decoding (referred to as the function \textit{Beam}) involves discrete argmax decisions and thus represents a discontinuous function.
\item The output, $\hat{y}$, of the \textit{Beam} function, which is the input to the loss function, $L(\hat{y},y^*)$, is discrete and hence the evaluation of the final loss is also discontinuous.
\end{enumerate}
We introduce a surrogate training objective that avoids these problems and as a result is fully continuous. In order to accomplish this, we propose a continuous relaxation to the \emph{composition} of our final loss metric, $L$, and our decoder function, $Beam$:
\[
\textit{softLB}(x, \mathcal{M}(\theta), y^*) \approx (L \circ Beam)(x, \mathcal{M}(\theta), y^*)
\]
Specifically, we form a continuous function \textit{softLB} that seeks to approximate the result of running our decoder on input $x$ and then evaluating the result against $y^*$ using $L$.
By introducing this new module, we are now able to construct our surrogate training objective:
\begin{align}
\min_\theta \tilde{G}_{\textrm{DL}}(x,\theta,y^*) = \min_\theta \textit{softLB}(x, \mathcal{M}(\theta), y^*)
\end{align}

Specified in more detail in Section~\ref{sec-soft-beam}, our surrogate objective in Equation 2 will additionally take a hyperparameter $\alpha$ that trades approximation quality for smoothness of the objective. Under certain conditions, Equation 2 converges to the objective in Equation 1 as $\alpha$ is increased. We first describe the standard discontinuous beam search procedure and then our training approach (Equation 2) involving a continuous relaxation of beam search.

\subsection{Discontinuity in Beam Search}
\begin{algorithm*}[t]
\caption{continuous-top-k-argmax}\label{peak}
\begin{algorithmic}[1]
\Inputs{$s \in \mathbb{R}^{k\times|V|}$}
\Outputs{$p_{i} \in \mathbb{R}^{k\times|V|}$, s.t. $\sum_j p_{ij}=1, i=1, \ldots, k$}
\State $m \in \mathbb{R}^{k} = \textit{top-k-max}(s)$
\For{$i$ = 1 to k} \Comment{\emph{peaked-softmax} will be dominated by scores closer to $m_i$}
\State $p_{i} = \textit{peaked-softmax}_{\alpha}(-(s - m_i\cdot{\mathbf{1}})^2)$ \Comment{The square operation is element-wise}
\EndFor
\end{algorithmic}
\end{algorithm*}
Formally, \textit{beam search} is a procedure with hyperparameter $k$ that maintains a beam of $k$ elements at each time step and expands each of the $k$ elements to find the $k$-best candidates for the next time step. The procedure finds an approximate argmax of a scoring function defined on output sequences.

We describe beam search in the context of seq2seq models in Algorithm~\ref{hard} -- more specifically, for an encoder-decoder \cite{sutskever2014sequence} model with a nonlinear auto-regressive decoder (e.g. an LSTM \cite{hochreiter1997long}). We define the global model score of a sequence $y$ with length $T$ to be the sum of local output scores at each time step of the seq2seq model: $s(y) = \sum_{t=1}^T f(h_t, y_t)$. In neural models, the function $f$ is implemented as a differentiable mapping, $\mathbb{R}^{|h|} \rightarrow \mathbb{R}^{|V|}$, which yields scores for vocabulary elements using the recurrent hidden states at corresponding time steps. In our notation, $h_{t,i}$ is the hidden state of the decoder at time step $t$ for beam element $i$, $e_{t,i}$ is the embedding of the output symbol at time-step $t$ for beam element $i$, and $s_{t,i}$ is the cumulative model score at step $t$ for beam element $i$. In Algorithm~\ref{hard}, we denote by $\tilde{s}_t \in \mathbb{R}^{k \times |V|}$ the cumulative candidate score matrix which represents the model score of each successor candidate in the vocabulary for each beam element. This score is obtained by adding the local output score (computed as $f(h_{t,i},w)$) to the running total of the score for the candidate. The function $r$ in Algorithms~\ref{hard} and~\ref{soft} yields successive hidden states in recurrent neural models like RNNs, LSTMs etc. The $\textit{embedding}$ operation maps a word in the vocabulary $V$, to a continuous embedding vector. Finally, backpointers at each time step to the beam elements at the previous time step are also stored for identifying the best sequence $\hat{y}$, at the conclusion of the search procedure. A backpointer at time step $t$ for a beam element $i$ is denoted by $\textit{b}_{t,i} \in \{1, \ldots, k\}$ which points to one of the $k$ elements at the previous beam. We denote a vector of backpointers for all the beam elements by $\textit{b}_{t,*}$. The $\textit{follow-backpointer}$ operation takes as input backpointers ($b_{t,*}$) and candidates ($y_{t,*} \in \{1, \ldots, |V|\}^k$) for all the beam elements at each time step and traverses the sequence in reverse (from time-step $T$ through 1) following backpointers at each time step and identifying candidate words associated with each backpointer that results in a sequence $\hat{y}$, of length $T$. 

The procedure described in Algorithm~\ref{hard} is discontinuous because of the \textit{top-k-argmax} procedure that returns a pair of vectors corresponding to the $k$ highest-scoring indices for backpointers and vocabulary items from the score matrix $\tilde s_t$. This index selection results in hard backpointers at each time step which restrict the gradient flow during backpropagation.
In the next section, we describe a continuous relaxation to the \textit{top-k-argmax} procedure which forms the crux of our approach.


\subsection{Continuous Approximation to \textit{top-k-argmax} \label{soft-topk}}

\begin{algorithm*}[t]
\caption{Continuous relaxation to beam search}\label{soft}
\begin{algorithmic}[1]
\Initialize{\strut$h_{0,i} \gets \vec{0}$, $e_{0,i} \gets \textit{embedding(<s>)}$, $s_{0,i} \gets 0$, $D_{t} \in \mathbb{R}^{k} \gets \vec{0}$, $i=1,\ldots,k$} 
\For{t = 0 to T}
\ForAll{$w \in V$}
\For{i=1 to k}
\State $\tilde{s}_t[i,w] \gets s_{t,i} + f(h_{t,i},w)$ \Comment{$f$ is a local output scoring function}
\EndFor
\State $\tilde{D}_{t,w} = d(w)$ \Comment{$\tilde{D}_t$ is used to compute $D_{t+1}$}
\EndFor
 \State $p_1,\ldots,p_k \gets \textit{continuous-top-k-argmax}(\tilde{s}_t)$ \Comment{Call Algorithm 2}
 \For{i = 1 to k}
  \State $\tilde{b}_{t,i} \gets \textit{row\_sum}(p_{i})$ \Comment{Soft back pointer computation}
  \State $a_i \in \mathcal{R}^{|V|} \gets \textit{column\_sum}(p_i)$ \Comment{Contribution from vocabulary items}
  \State $e_{t+1,i} \gets a_i^{T} \times \textit E$ \Comment{Peaked distribution over the candidates to compute $e,D,S$}
  \State $D_{t+1,i} \gets a_i^{T} \cdot \tilde{D}_t$
  \State $s_{t+1,i} = \textit{sum}(\tilde{s}_t \odot p_i)$
  \State $\tilde{h}_{t,i} \gets \vec{0}$
  \For{j = 1 to k} \Comment{Get contributions from soft backpointers for each beam element}
  \State $\tilde{h}_{t,i} += h_{t,j}*\tilde{b}_{t,i}[j]$
  \State $D_{t+1,i} += D_{t,j}*\tilde{b}_{t,i}[j]$
  \EndFor
  \State $h_{t+1,i} \gets r(\tilde{h}_{t,i},e_{t+1,i})$ \Comment{$r$ is a nonlinear recurrent function that returns state at next step}
 \EndFor
\EndFor
\State $L = \textit{peaked-softmax}_{\alpha}(s_T) \cdot D_{T}$ \Comment{Pick the loss for the sequence with highest model score on the beam in a soft manner.}
\end{algorithmic}
\end{algorithm*}
\begin{figure*}[t]
\centering
\includegraphics[height=8.5cm]{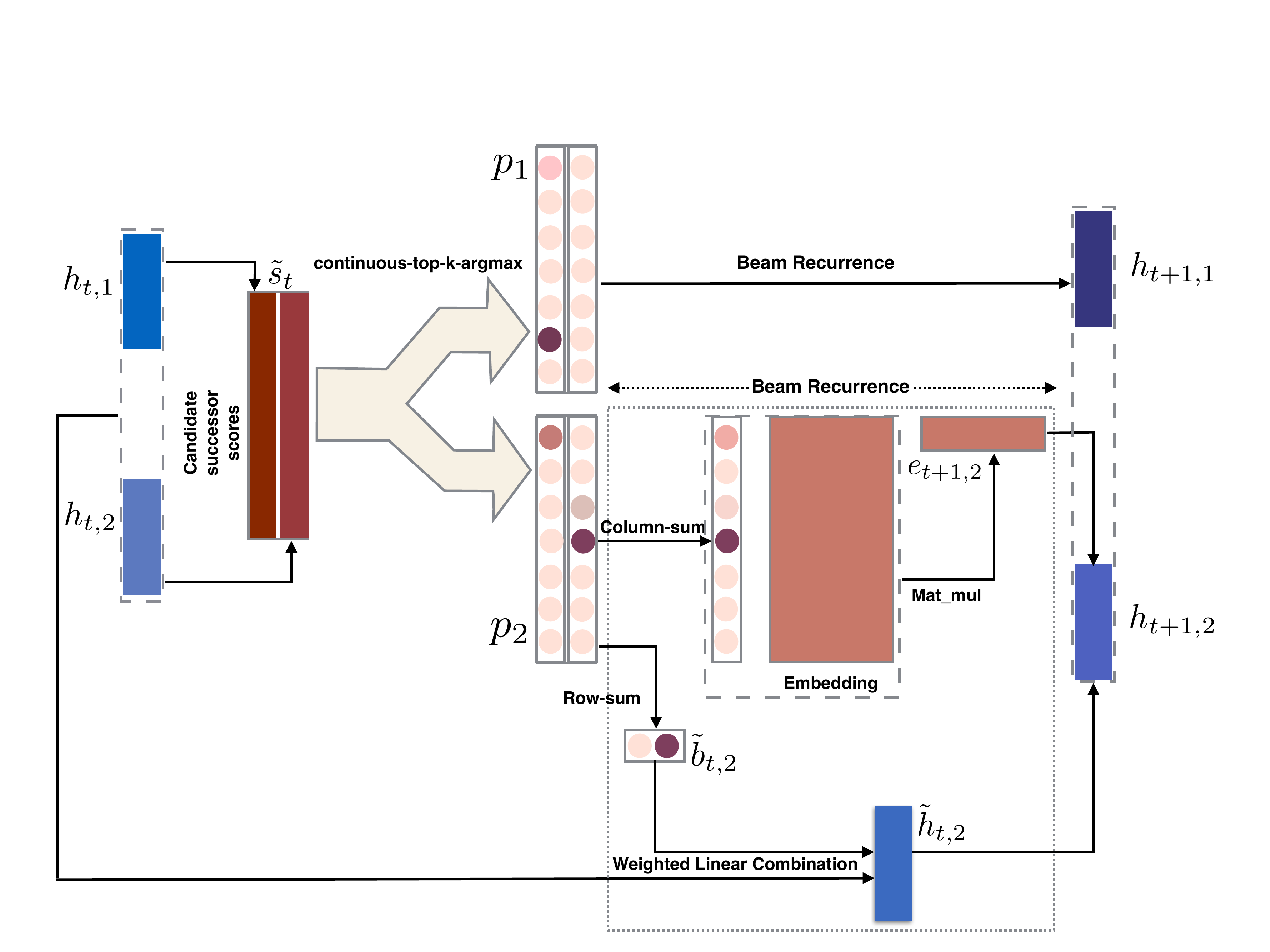}
  \label{figmodel}
  \caption{Illustration of our approximate continuous beam search (Algorithm~\ref{soft}) module to obtain hidden states for beam elements at the next time step (${h_{t+1,*}}$), starting from the hidden states corresponding to beam elements are current time step ($h_{t,*}$) with beam size of 2. `Beam recurrence' module has been expanded for $h_{t+1,2}$ and similar procedure is carried out for $h_{t+1,1}$.}
\end{figure*}
The key property that we use in our approximation is that for a real valued vector $z$, the argmax with respect to a vector of scores, $s$, can be approximated by a temperature controlled softmax operation. The argmax operation can be represented as:
\begin{align*}
    \hat{z} = \sum_i z_i\mathbbm{1}[\forall i' \ne i,\ \ s_i > s_{i'}],
\end{align*}
which can be relaxed by replacing the indicator function with a \emph{peaked-softmax} operation with hyperparameter $\alpha$:
\begin{align*}
 \begin{split}
    \tilde{z} = \sum_i z_i~ \frac{\exp{(\alpha~ s_i})}{\sum_{i'} \exp{(\alpha~ s_{i'})}} = & z^T \cdot \frac{\textrm{elem-exp}(\alpha~ s)}{\sum_{i'} \exp{(\alpha~ s_{i'})}}\\= z^T \cdot \textit{peaked-softmax}_{\alpha}(\mathbf{s})
 \end{split}
\end{align*}
As $\alpha \to \infty$, $\tilde{z} \to \hat{z}$ so long as there is only one maximum value in the vector $z$. This \emph{peaked-softmax} operation has been shown to be effective in recent work \cite{maddison2016concrete,jang2016categorical,softgreedy} involving continuous relaxation to the argmax operation, although to our knowledge, this is the first work to apply it to approximate the beam search procedure. 

Using this \emph{peaked-softmax} operation, we propose an iterative algorithm for computing a continuous relaxation to the \textit{top-k-argmax} procedure in Algorithm~\ref{peak} which takes as input a score matrix of size $k\times |V|$ and returns $k$ peaked matrices $p$ of size $k \times |V|$. Each matrix $p_{i}$ represents the index of $i$-th max. For example, $p_1$ will have most of its mass concentrated on the index in the matrix that corresponds to the argmax, while $p_2$ will have most of its mass concentrated on the index of the 2nd-highest scoring element. Specifically, we obtain matrix $p_i$ by computing the squared difference between the $i$-highest score and all the scores in the matrix and then using the \emph{peaked-softmax} operation over the negative squared differences. This results in scores closer to the $i$-highest score to have a higher mass than scores far away from the $i$-highest score.

Hence, the continuous relaxation to \textit{top-k-argmax} operation can be simply implemented by iteratively using the \textit{max} operation which is continuous and allows for gradient flow during backpropagation. As $\alpha \to \infty$, each $p$ vector converges to hard index pairs representing hard backpointers and successor candidates described in Algorithm~\ref{hard}. For finite $\alpha$, we introduce a notion of a soft backpointer, represented as a vector $\tilde{b} \in \mathbb{R}^k$ in the $k$-probability simplex, which represents the contribution of each beam element from the previous time step to a beam element at current time step. This is obtained by a row-wise sum over $p$ to get $k$ values representing soft backpointers. 

\subsection{Training with Continuous Relaxation of Beam Search \label{sec-soft-beam}}
We describe our approach in detail in Algorithm~3 and illustrate the soft beam recurrence step in Figure~1. For composing the loss function and the beam search function for our optimization as proposed in Equation 2, we make use of decomposability of the loss function across time-steps. Thus for a sequence y, the total loss is: $L(y,y^*) = \sum_{t=1}^T d(y_t)$. In our experiments, $d(y_t)$ is the Hamming loss which can be easily computed at each time-step by simply comparing gold $y_t^*$ with $y_t$. While exact computation of $d(y)$ will vary according to the loss, our proposed procedure will be applicable as long as the total loss is decomposable across time-steps. While decomposability of loss is a strong assumption, existing literature on structured prediction \cite{taskar2004max,tsochantaridis2005large} has made due with this assumption, often using decomposable losses as surrogates for non-decomposable ones. We detail the continuous relaxation to beam search in Algorithm~\ref{soft} with $D_{t,i}$ being the cumulative loss of beam element $i$ at time step $t$ and $E$ being the embedding matrix of the target vocabulary which is of size $|V|\times l$ where $l$ is the size of the embedding vector.

In Algorithm~\ref{soft}, all the discrete selection functions have been replaced by their soft, continuous counterparts which can be backpropagated through. This results in all the operations being matrix and vector operations which is ideal for a GPU implementation. An important aspect of this algorithm is that we no longer rely on exactly identifying a discrete search prediction $\hat{y}$ since we are only interested in a continuous approximation to the direct loss $L$ (line 18 of Algorithm~\ref{soft}), and all the computation is expressed via the \emph{soft beam search} formulation which eliminates all the sources of discontinuities associated with the training objective in Equation 1. The computational complexity of our approach for training scales linearly with the beam size and hence is roughly $k$ times slower than standard CE training for beam size $k$.
Since we have established the pointwise convergence of \emph{peaked-softmax} to argmax as $\alpha \to \infty$ for all vectors that have a unique maximum value, we can establish pointwise convergence of objective in Equation 2 to objective in Equation 1 as $\alpha \to \infty$, as long as there are no ties among the top-k scores of the beam expansion candidates at any time step. We posit that absolute ties are unlikely due to random initialization of weights and the domain of the scores being $\mathbb{R}$. Empirically, we did not observe any noticeable impact of potential ties on the training procedure and our approach performed well on the tasks as discussed in Section~\ref{results}.
\begin{align}
\tilde{G}_{\textrm{DL},\alpha}(x, \mathcal{M}(\theta), y^*) \xrightarrow[\textit{p}]{\alpha \to \infty} G_{\textrm{DL}}(x,\theta, y^*)
\end{align}
We experimented with different annealing schedules for $\alpha$ starting with \emph{non-peaked softmax} moving toward \emph{peaked-softmax} across epochs so that learning is stable with informative gradients. This is important because cost functions like Hamming distance with very high $\alpha$ tend to be non-smooth and are generally flat in regions far away from changepoints and have a very large gradient near the changepoints which makes optimization difficult.

\subsection{Decoding \label{decoding}}
The motivation behind our approach is to make the optimization aware of beam search decoding while maintaining the continuity of the objective. However, since our approach doesn't introduce any new model parameters and optimization is agnostic to the architecture of the seq2seq model, we were able to experiment with various decoding schemes like locally normalized greedy decoding, and hard beam search, once the model has been trained.

However, to reduce the gap between the training procedure and test procedure, we also experimented with \emph{soft beam search decoding}. This decoding approach closely follows Algorithm~\ref{soft}, but along with soft back pointers, we also compute hard back pointers at each time step. After computing all the relevant quantities like model score, loss etc., we follow the \emph{hard} backpointers to obtain the best sequence $\hat{y}$. This is very different from hard beam decoding because at each time step, the selection decisions are made via our soft continuous relaxation which influences the scores, LSTM hidden states and input embeddings at subsequent time-steps. The hard backpointers are essentially the MAP estimate of the soft backpointers at each step. With small, finite $\alpha$, we observe differences between soft beam search and hard beam search decoding in our experiments.

\subsection{Comparison with Max-Margin Objectives \label{maxm}}
Max-margin based objectives are typically motivated as another kind of surrogate training objective which avoid the discontinuities associated with direct loss optimization. Hinge loss for structured prediction typically takes the form:
\begin{align*}
    G_{\textit{hinge}} = \max(0,\max_{y \in \mathcal{Y}}(\Delta(y,y^*)+ s(y)) - s(y^*))
\end{align*}
where $x$ is the input sequence, $y^*$ is the gold target sequence, $\mathcal{Y}$ is the output search space and $\Delta(y,y^*)$ is the discontinuous cost function which we assume is decomposable across the time-steps of a sequence. Finding the cost augmented maximum score is generally difficult in large structured models and often involves searching over the output space and computing the approximate cost augmented maximal output sequence and the score associated with it via beam search. This procedure introduces discontinuities in the training procedure of structured max-margin objectives and renders it non amenable to training via backpropagation. Related work \cite{wiseman2016sequence} on incorporating beam search into the training of neural sequence models does involve cost-augmented max-margin loss but it relies on discontinuous beam search forward passes and an explicit mechanism to ensure that the gold sequence stays in the beam during training, and hence does not involve back propagation through the beam search procedure itself.

Our continuous approximation to beam search can very easily be modified to compute an approximation to the structured hinge loss so that it can be trained via backpropagation if the cost function is decomposable across time-steps. In Algorithm~\ref{soft}, we only need to modify line 5 as:
\begin{align*}
    \tilde{s}_t[i,w] \gets s_{t,i} + d(w) + f(h_{t,i},w)
\end{align*}
and instead of computing $L$ in Algorithm~\ref{soft}, we first compute the cost augmented maximum score as:
\begin{align*}
  s_{\textit{max}} = \textit{peaked-softmax}_{\alpha}(s_T) \cdot s_{T}
\end{align*}
and also compute the target score $s(y^*)$ by simply running the forward pass of the LSTM decoder over the gold target sequence. The continuous approximation to the hinge loss to be optimized is then: $\tilde{G}_{\textrm{hinge},\alpha} = \max(0,s_{\textit{max}}-s(y^*))$. We empirically compare this approach with the proposed approach to optimize direct loss in experiments.
\section{Experimental Setup}
Since our goal is to investigate the efficacy of our approach for training generic seq2seq models, we perform experiments on two NLP tagging tasks with very different characteristics and output search spaces: Named Entity Recognition (NER) and CCG supertagging. While seq2seq models are appropriate for CCG supertagging task because of the long-range correlations between the sequential output elements and a large search space, they are not ideal for NER which has a considerably smaller search space and weaker correlations between predictions at subsequent time steps. In our experiments, we observe improvements from our approach on \emph{both} of the tasks.  
We use a seq2seq model with a bi-directional LSTM encoder (1 layer with tanh activation function) for the input sequence $x$, and an LSTM decoder (1 layer with tanh activation function) with a fixed attention mechanism that deterministically attends to the $i$-th input token when decoding the $i$-th output, and hence does not involve learning of any attention parameters. Since, computational complexity of our approach for optimization scales linearly with beam size for each instance, it is impractical to use very large beam sizes for training. Hence, beam size for all the beam search based experiments was set to 3 which resulted in improvements on both the tasks as discussed in the results. For both tasks, the direct loss function was the \textit{Hamming distance cost} which aims to maximize word level accuracy.
\subsection{Named Entity Recognition}
For named entity recognition, we use the CONLL 2003 shared task data \cite{tjong2003introduction} for German language and use the provided data splits. We perform no preprocessing on the data. The output vocabulary length (label space) is 10. A peculiar characteristic of this problem is that the training data is naturally skewed toward one default label (`O') because sentences typically do not contain many named entities and the evaluation focuses on the performance recognizing entities. Therefore, we modify the Hamming cost such that incorrect prediction of `O' is doubly penalized compared to other incorrect predictions. We use the hidden layers of size 64 and label embeddings of size 8. As mentioned earlier, seq2seq models are not an ideal choice for NER (tag-level correlations are short-ranged in NER -- the unnecessary expressivity of full seq2seq models over simple encoder-classifier neural models makes training harder). However, we wanted to evaluate the effectiveness of our approach on different instantiations of seq2seq models.
\subsection{CCG Supertagging}
We used the standard splits of CCG bank \cite{hockenmaier2002acquiring} for training, development, and testing. The label space of supertags is 1,284 which is much larger than NER. The distribution of supertags in the training data exhibits a long tail because these supertags encode specific syntactic information about the words' usage. The supertag labels are correlated with each other and many tags encode similar information about the syntax. Moreover, this task is sensitive to the long range sequential decisions and search effects because of how it holistically encodes the syntax of the entire sentence. We perform minor preprocessing on the data similar to the preprocessing in \cite{vaswani2016supertagging}. For this task, we used hidden layers of size 512 and the supertag label embeddings were also of size 512. The standard evaluation metric for this task is the word level label accuracy which directly corresponds to Hamming loss.
\subsection{Hyperparameter tuning}
For tuning all the hyperparameters related to optimization we trained our models for 50 epochs and picked the models with the best performance on the development set. We also ran multiple random restarts for all the systems evaluated to account for performance variance across randomly started runs. We pretrained all our models with standard cross entropy training which was important for stable optimization of the non convex neural objective with a large parameter search space. This warm starting is a common practice in prior work on complex neural models \cite{ranzato2015sequence,rush2015neural,bengioss}. 
\hspace{-0.2cm}\begin{table*}[t]
\centering
\caption{Results on CCG Supertagging. Tag-level accuracy is reported in this table which is a standard evaluation metric for supertagging}
\label{super}
\scalebox{1.0}{
\begin{tabular}{lllllll}
\toprule
Training procedure & \multicolumn{2}{c}{Greedy} & \multicolumn{2}{c}{Hard Beam Search} & \multicolumn{2}{c}{Soft Beam Search} \\
& Dev & Test & Dev & Test & Dev & Test\\
\midrule
\midrule 
Baseline CE     &  80.15 & 80.35 & 82.17   &  82.42 &  81.62 &  82.00 \\
\midrule
     
$\tilde{G}_{\textrm{hinge},\alpha}$ annealed $\alpha$   & -        & -       & 83.03        & 83.54      &    82.82     & 83.05\\        
$\tilde{G}_{\textrm{hinge},\alpha} \alpha$=1.0 & -        & -       & 83.02        & 83.36      &    82.49     & 82.85\\       
$\tilde{G}_{\textrm{DL},\alpha} \alpha$=1.0 & -        & -       & 83.23        & 82.65      &    82.58     & 82.82\\  
$\tilde{G}_{\textrm{DL},\alpha}$ annealed $\alpha$   & -        & -       & \bf 85.69        & \bf 85.82     &    85.58     & 85.78\\  
\bottomrule
\end{tabular}
}
\end{table*}
\hspace{-0.2cm}\begin{table*}[t]
\centering
\caption{Results on Named Entity Recognition. Macro F1 over the prediction of different named entities is reported that is a standard evaluation metric for this task.}
\label{NER}
\scalebox{1.0}{
\begin{tabular}{lllllll}
\toprule
Training procedure & \multicolumn{2}{c}{CE Greedy} & \multicolumn{2}{c}{Hard Beam Search} & \multicolumn{2}{c}{Soft Beam Search} \\
& Dev & Test & Dev & Test & Dev & Test\\
\midrule
\midrule 
Baseline CE     &  50.21 & 54.92 & 46.22   &  51.34 &  47.50 &  52.78 \\
\midrule
$\tilde{G}_{\textrm{hinge},\alpha}$ annealed $\alpha$   & -        & -       & 41.10        & 45.98      &    41.24     & 46.34\\        
$\tilde{G}_{\textrm{hinge},\alpha} \alpha$=1.0 & -        & -       & 40.09        & 44.67      &   39.67      & 43.82\\          
$\tilde{G}_{\textrm{DL},\alpha} \alpha$=1.0 & -        & -       & 49.88        & 54.08      &    50.73     & 54.77\\  
$\tilde{G}_{\textrm{DL},\alpha}$ annealed $\alpha$   & -        & -       &  51.86        & 56.15      &    \bf 51.96     & \bf 56.38\\ 
\bottomrule
\end{tabular}
}
\end{table*}
\subsection{Comparison}
We report performance on validation and test sets for both the tasks in Tables~1 and ~2. The baseline model is a cross entropy trained seq2seq model (Baseline CE) which is also used to warm start the the proposed optimization procedures in this paper. This baseline has been compared against the approximate direct loss training objective (Section~\ref{sec-soft-beam}), referred to as $\tilde{G}_{\textrm{DL},\alpha}$ in the tables, and the approximate max-margin training objective (Section~\ref{maxm}), referred to as $\tilde{G}_{\textrm{hinge},\alpha}$ in the tables. Results are reported for models when trained with annealing $\alpha$, and also with a constant setting of $\alpha=1.0$ which is a very smooth but inaccurate approximation of the original direct loss that we aim to optimize\footnote{Our pilot experiments that involved training with a very large constant $\alpha$ resulted in unstable optimization.}. Comparisons have been made on the basis of performance of the models under different decoding paradigms (represented as different column in the tables): locally normalized decoding (CE greedy), hard beam search decoding and soft beam search decoding described in Section~\ref{decoding}.
\section{Results}
\label{results}
As shown in Tables~1 and 2, our approach $\tilde{G}_{\textrm{DL},\alpha}$ shows significant improvements over the locally normalized CE baseline with greedy decoding for both the tasks (+5.5 accuracy points gain for supertagging and +1.5 F1 points for NER). The improvement is more pronounced on the supertagging task, which is not surprising because: (i) the evaluation metric is tag-level accuracy which is congruent with the Hamming loss that $\tilde{G}_{\textrm{DL},\alpha}$ directly optimizes and (ii) the supertagging task itself is very sensitive to the search procedure because tags across time-steps tend to exhibit long range dependencies as they encode specialized syntactic information about word usage in the sentence. 

Another common trend to observe is that annealing $\alpha$ always results in better performance than training with a constant $\alpha=1.0$ for both $\tilde{G}_{\textrm{DL},\alpha}$ (Section~\ref{sec-soft-beam}) and $\tilde{G}_{\textrm{hinge},\alpha}$ (Section~\ref{maxm}). This shows that a stable training scheme that smoothly approaches minimizing the actual direct loss is important for our proposed approach. Additionally, we did not observe a large difference when our soft approximation is used for decoding (Section~\ref{decoding}) compared to hard beam search decoding, which suggests that our approximation to the hard beam search is as effective as its discrete counterpart.

For supertagging, we observe that the baseline cross entropy trained model improves its predictions with  beam search decoding compared to greedy decoding by 2 accuracy points, which suggests that beam search is already helpful for this task, even without search-aware training. Both the optimization schemes proposed in this paper improve upon the baseline with soft direct loss optimization ($\tilde{G}_{\textrm{DL},\alpha}$), performing better than the approximate max-margin approach. \footnote{Separately, we also ran experiments with a max-margin objective that used hard beam search to compute loss-augmented decodes. This objective is discontinuous, but we evaluated the performance of gradient optimization nonetheless. While not included in the result tables, we found that this approach was unstable and considerably underperformed both approximate max-margin 
and 
direct loss objectives.} 

For NER, we observe that optimizing $\tilde{G}_{\textrm{DL},\alpha}$ outperforms all the other approaches but we also observe interesting behaviour of beam search decoding and the approximate max-margin objective for this task. The pretrained CE baseline model yields worse performance when beam search is done instead of greedy locally normalized decoding. This is because the training data is heavily skewed toward the `O' label and hence the absolute score resolution between different tags at each time-step during decoding isn't enough to avoid leading beam search toward a wrong hypothesis path. We observed in our experiments that hard beam search resulted in predicting more `O's which also hurt the prediction of tags at future time steps and hurt precision as well as recall. Encouragingly, $\tilde{G}_{\textrm{DL},\alpha}$ optimization, even though warm started with a CE trained model that performs worse with beam search, led to the NER model becoming more search aware, which resulted in superior performance. However, we also observe that the approximate max-margin approach ($\tilde{G}_{\textrm{hinge},\alpha}$) performs poorly here. We attribute this to a deficiency in the max-margin objective when coupled with approximate search methods like beam search that do not provide guarantees on finding the supremum: one way to drive this objective down is to learn model scores such that the search for the best hypothesis is difficult, so that the value of the loss augmented decode is low, while the gold sequence maintains higher model score. Because we also warm started with a pre-trained model that results in a worse performance with beam search decode than with greedy decode, we observe the adverse effect of this deficiency. The result is a model that scores the gold hypothesis highly, but yields poor decoding outputs. This observation indicates that using max-margin based objectives with beam search during \emph{training} actually may achieve the opposite of our original intent: the objective can be driven down by \emph{introducing} search errors.

The observation that our optimization method led to improvements on \emph{both} the tasks--even on NER for which hard beam search during decoding on a CE trained model hurt the performance--by making the optimization more search aware, indicates the effectiveness of our approach for training seq2seq models.
\section{Conclusion}
While beam search is a method of choice for performing search in neural sequence models, as our experiments confirm, it is not necessarily guaranteed to improve accuracy when applied to cross-entropy-trained models.
In this paper, we propose a novel method for optimizing model parameters that directly takes into account the process of beam search itself through a continuous, end-to-end sub-differentiable relaxation of beam search composed with the final evaluation loss.
Experiments demonstrate that our method is able to improve overall test-time results for models using beam search as a test-time inference method, leading to substantial improvements in accuracy.

\bibliography{refs.bib}

\begin{thebibliography}{}

\bibitem[\protect\citeauthoryear{Andor \bgroup et al\mbox.\egroup
  }{2016}]{andor2016globally}
Andor, D.; Alberti, C.; Weiss, D.; Severyn, A.; Presta, A.; Ganchev, K.;
  Petrov, S.; and Collins, M.
\newblock 2016.
\newblock Globally normalized transition-based neural networks.
\newblock In {\em Association for Computational Linguistics}.

\bibitem[\protect\citeauthoryear{Bahdanau \bgroup et al\mbox.\egroup
  }{2016}]{bahdanau2016task}
Bahdanau, D.; Serdyuk, D.; Brakel, P.; Ke, N.~R.; Chorowski, J.; Courville, A.;
  and Bengio, Y.
\newblock 2016.
\newblock Task loss estimation for structured prediction.

\bibitem[\protect\citeauthoryear{Bahdanau \bgroup et al\mbox.\egroup
  }{2017}]{bahdanau2016actor}
Bahdanau, D.; Brakel, P.; Xu, K.; Goyal, A.; Lowe, R.; Pineau, J.; Courville,
  A.; and Bengio, Y.
\newblock 2017.
\newblock An actor-critic algorithm for sequence prediction.
\newblock In {\em International Conference on Learning Representations}.

\bibitem[\protect\citeauthoryear{Bahdanau, Cho, and
  Bengio}{2015}]{bahdanau2014neural}
Bahdanau, D.; Cho, K.; and Bengio, Y.
\newblock 2015.
\newblock Neural machine translation by jointly learning to align and
  translate.
\newblock In {\em International Conference on Learning Representations}.

\bibitem[\protect\citeauthoryear{Bengio \bgroup et al\mbox.\egroup
  }{2015}]{bengioss}
Bengio, S.; Vinyals, O.; Jaitly, N.; and Shazeer, N.
\newblock 2015.
\newblock Scheduled sampling for sequence prediction with recurrent neural
  networks.
\newblock In {\em Advances in Neural Information Processing Systems},
  1171--1179.

\bibitem[\protect\citeauthoryear{Cho \bgroup et al\mbox.\egroup
  }{2014}]{cho2014properties}
Cho, K.; Van~Merri{\"e}nboer, B.; Bahdanau, D.; and Bengio, Y.
\newblock 2014.
\newblock On the properties of neural machine translation: Encoder-decoder
  approaches.
\newblock {\em arXiv preprint arXiv:1409.1259}.

\bibitem[\protect\citeauthoryear{Collins and
  Roark}{2004}]{collins2004incremental}
Collins, M., and Roark, B.
\newblock 2004.
\newblock Incremental parsing with the perceptron algorithm.
\newblock In {\em Proceedings of the 42nd Annual Meeting on Association for
  Computational Linguistics},  111.
\newblock Association for Computational Linguistics.

\bibitem[\protect\citeauthoryear{Daum{\'e}~III and
  Marcu}{2005}]{daume2005learning}
Daum{\'e}~III, H., and Marcu, D.
\newblock 2005.
\newblock Learning as search optimization: Approximate large margin methods for
  structured prediction.
\newblock In {\em Proceedings of the 22nd international conference on Machine
  learning},  169--176.
\newblock ACM.

\bibitem[\protect\citeauthoryear{Daum{\'e}, Langford, and
  Marcu}{2009}]{daume2009search}
Daum{\'e}, H.; Langford, J.; and Marcu, D.
\newblock 2009.
\newblock Search-based structured prediction.
\newblock {\em Machine learning} 75(3):297--325.

\bibitem[\protect\citeauthoryear{Dyer \bgroup et al\mbox.\egroup
  }{2015}]{dyer2015transition}
Dyer, C.; Ballesteros, M.; Ling, W.; Matthews, A.; and Smith, N.~A.
\newblock 2015.
\newblock Transition-based dependency parsing with stack long short-term
  memory.
\newblock {\em arXiv preprint arXiv:1505.08075}.

\bibitem[\protect\citeauthoryear{Dyer \bgroup et al\mbox.\egroup
  }{2016}]{dyer2016recurrent}
Dyer, C.; Kuncoro, A.; Ballesteros, M.; and Smith, N.~A.
\newblock 2016.
\newblock Recurrent neural network grammars.
\newblock In {\em Proceedings of NAACL-HLT},  199--209.

\bibitem[\protect\citeauthoryear{Gormley, Dredze, and
  Eisner}{2015}]{gormley2015approximation}
Gormley, M.~R.; Dredze, M.; and Eisner, J.
\newblock 2015.
\newblock Approximation-aware dependency parsing by belief propagation.
\newblock {\em Transactions of the Association for Computational Linguistics
  (TACL)}.

\bibitem[\protect\citeauthoryear{Goyal, Dyer, and
  Berg-Kirkpatrick}{2017}]{softgreedy}
Goyal, K.; Dyer, C.; and Berg-Kirkpatrick, T.
\newblock 2017.
\newblock Differentiable scheduled sampling for credit assignment.
\newblock In {\em Association for Computational Linguistics}.

\bibitem[\protect\citeauthoryear{Gu, Im, and Li}{2017}]{gu2017neural}
Gu, J.; Im, D.~J.; and Li, V.~O.
\newblock 2017.
\newblock Neural machine translation with gumbel-greedy decoding.
\newblock {\em arXiv preprint arXiv:1706.07518}.

\bibitem[\protect\citeauthoryear{Hochreiter and
  Schmidhuber}{1997}]{hochreiter1997long}
Hochreiter, S., and Schmidhuber, J.
\newblock 1997.
\newblock Long short-term memory.
\newblock {\em Neural computation} 9(8):1735--1780.

\bibitem[\protect\citeauthoryear{Hockenmaier and
  Steedman}{2002}]{hockenmaier2002acquiring}
Hockenmaier, J., and Steedman, M.
\newblock 2002.
\newblock Acquiring compact lexicalized grammars from a cleaner treebank.
\newblock In {\em LREC}.

\bibitem[\protect\citeauthoryear{Jang, Gu, and
  Poole}{2016}]{jang2016categorical}
Jang, E.; Gu, S.; and Poole, B.
\newblock 2016.
\newblock Categorical reparameterization with gumbel-softmax.
\newblock In {\em International Conference on Learning Representations}.

\bibitem[\protect\citeauthoryear{Koehn and Knowles}{2017}]{koehn2017six}
Koehn, P., and Knowles, R.
\newblock 2017.
\newblock Six challenges for neural machine translation.
\newblock {\em arXiv preprint arXiv:1706.03872}.

\bibitem[\protect\citeauthoryear{Maddison, Mnih, and
  Teh}{2017}]{maddison2016concrete}
Maddison, C.~J.; Mnih, A.; and Teh, Y.~W.
\newblock 2017.
\newblock The concrete distribution: A continuous relaxation of discrete random
  variables.
\newblock In {\em International Conference on Learning Representations}.

\bibitem[\protect\citeauthoryear{Neubig}{2017}]{neubig2017neural}
Neubig, G.
\newblock 2017.
\newblock Neural machine translation and sequence-to-sequence models: A
  tutorial.
\newblock {\em arXiv preprint arXiv:1703.01619}.

\bibitem[\protect\citeauthoryear{Ranzato \bgroup et al\mbox.\egroup
  }{2016}]{ranzato2015sequence}
Ranzato, M.; Chopra, S.; Auli, M.; and Zaremba, W.
\newblock 2016.
\newblock Sequence level training with recurrent neural networks.
\newblock In {\em International Conference on Learning Representations}.

\bibitem[\protect\citeauthoryear{Ross, Gordon, and
  Bagnell}{2011}]{ross2011reduction}
Ross, S.; Gordon, G.~J.; and Bagnell, D.
\newblock 2011.
\newblock A reduction of imitation learning and structured prediction to
  no-regret online learning.
\newblock In {\em AISTATS}, volume~1, ~6.

\bibitem[\protect\citeauthoryear{Rush, Chopra, and
  Weston}{2015}]{rush2015neural}
Rush, A.~M.; Chopra, S.; and Weston, J.
\newblock 2015.
\newblock A neural attention model for abstractive sentence summarization.
\newblock In {\em Empirical Methods in Natural Language Processing}.

\bibitem[\protect\citeauthoryear{Serban \bgroup et al\mbox.\egroup
  }{2015}]{serban2015building}
Serban, I.~V.; Sordoni, A.; Bengio, Y.; Courville, A.; and Pineau, J.
\newblock 2015.
\newblock Building end-to-end dialogue systems using generative hierarchical
  neural network models.
\newblock In {\em AAAI'16 Proceedings of the Thirtieth AAAI Conference on
  Artificial Intelligence}.

\bibitem[\protect\citeauthoryear{Sutskever, Vinyals, and
  Le}{2014}]{sutskever2014sequence}
Sutskever, I.; Vinyals, O.; and Le, Q.~V.
\newblock 2014.
\newblock Sequence to sequence learning with neural networks.
\newblock In {\em Advances in neural information processing systems},
  3104--3112.

\bibitem[\protect\citeauthoryear{Taskar, Guestrin, and
  Koller}{2004}]{taskar2004max}
Taskar, B.; Guestrin, C.; and Koller, D.
\newblock 2004.
\newblock Max-margin markov networks.
\newblock In {\em Advances in neural information processing systems},  25--32.

\bibitem[\protect\citeauthoryear{Tjong Kim~Sang and
  De~Meulder}{2003}]{tjong2003introduction}
Tjong Kim~Sang, E.~F., and De~Meulder, F.
\newblock 2003.
\newblock Introduction to the conll-2003 shared task: Language-independent
  named entity recognition.
\newblock In {\em Proceedings of the seventh conference on Natural language
  learning at HLT-NAACL 2003-Volume 4},  142--147.
\newblock Association for Computational Linguistics.

\bibitem[\protect\citeauthoryear{Tsochantaridis \bgroup et al\mbox.\egroup
  }{2005}]{tsochantaridis2005large}
Tsochantaridis, I.; Joachims, T.; Hofmann, T.; and Altun, Y.
\newblock 2005.
\newblock Large margin methods for structured and interdependent output
  variables.
\newblock {\em Journal of machine learning research} 6(Sep):1453--1484.

\bibitem[\protect\citeauthoryear{Vaswani \bgroup et al\mbox.\egroup
  }{2016}]{vaswani2016supertagging}
Vaswani, A.; Bisk, Y.; Sagae, K.; and Musa, R.
\newblock 2016.
\newblock Supertagging with lstms.
\newblock In {\em Proceedings of NAACL-HLT},  232--237.

\bibitem[\protect\citeauthoryear{Wiseman and Rush}{2016}]{wiseman2016sequence}
Wiseman, S., and Rush, A.~M.
\newblock 2016.
\newblock Sequence-to-sequence learning as beam-search optimization.
\newblock In {\em Empirical Methods in Natural Language Processing}.

\bibitem[\protect\citeauthoryear{Xu \bgroup et al\mbox.\egroup
  }{2015}]{xu2015show}
Xu, K.; Ba, J.; Kiros, R.; Cho, K.; Courville, A.~C.; Salakhutdinov, R.; Zemel,
  R.~S.; and Bengio, Y.
\newblock 2015.
\newblock Show, attend and tell: Neural image caption generation with visual
  attention.
\newblock In {\em ICML}, volume~14,  77--81.

\end{thebibliography}
\bibliographystyle{aaai}
\end{document}